\documentclass{article}
\usepackage{arxiv}
\usepackage[utf8]{inputenc} 
\usepackage[T1]{fontenc}    
\usepackage{hyperref}       
\usepackage{url}            
\usepackage{booktabs}       
\usepackage{amsmath,amsfonts}       
\usepackage{nicefrac}       
\usepackage{microtype}      
\usepackage{lipsum}		
\usepackage{graphicx}
\usepackage{natbib}
\usepackage{tabularx}
\usepackage{longtable}
\usepackage{doi}
\usepackage{algorithm}
\usepackage{algpseudocode}
\usepackage{array}
\usepackage{amssymb}
\usepackage{natbib}

\newcolumntype{R}[1]{>{\raggedleft\arraybackslash}m{#1}}  

\newcolumntype{L}[1]{>{\raggedright\arraybackslash}m{#1}}

\newcolumntype{C}[1]{>{\centering\arraybackslash}m{#1}}

\title{Spectral-Morphological Attention U-Net: An Efficient Network for Active Wildfire Detection}

\author{ \href{https://orcid.org/0009-0005-1489-5911}{\includegraphics[scale=0.06]{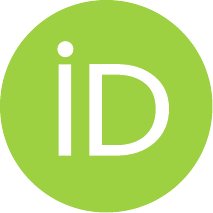}\hspace{1mm}Yugong Zeng}\\
	Department of Electrical and Computer Engineering\\
	University of Windsor\\
    401 Sunset Avenue\\
	Ontario, Canada N9B 3P4 \\
	\texttt{zeng26@uwindsor.ca} \\
	\And
	{\hspace{1mm}Jonathan Wu} \\
	Department of Electrical and Computer Engineering\\
	University of Windsor\\
	401 Sunset Avenue \\
    Ontario, Canada N9B 3P4 \\
	\texttt{jwu@uwindsor.ca} \\
}

\date{}

\renewcommand{\shorttitle}{\textit{}}

\begin{document}
\maketitle

\begin{abstract}
Over the past decades, the frequency of global wildfires has been increasing steadily. Therefore, if the fire can be detected and precisely located at an early stage, the potential hazards caused by it can be minimized to the greatest extent. The machine learning methods based on satellite images, due to their ability to automatically monitor extremely remote and vast areas, have shown great potential for application in the field of wildfire detection. To address this challenge, we proposed a new model named spectral-morphological attention U-Net(SMA-UNet), which includes a spectral attention module, a residual attention UNet backbone, a channel-spatial modulator, and a pair of differentiable morphological gates. We trained and evaluated this model with two datasets. These modules, excluding the backbone, are used to detect active fire events for the first time, especially the pair of differentiable morphological gates, which is innovatively developed. The proposed model achieved the highest scores in both datasets (e.g., intersection over union 75.16\% in TS-SatFire, 22.50\% in Sen2Fire). By conducting ablation studies of each module, we compared their independent contributions and tested their combinations. Ultimately, the integration of these modules yields a highly robust framework that significantly improves segmentation consistency across diverse and complex environmental conditions. Future work will focus on validating the proposed architecture across large-scale, multi-regional datasets from different satellite sensors to establish its broader generalizability for global wildfire detection.

\end{abstract}


\section{Introduction}
Wildfires represent one of the most critical natural hazards, posing severe threats to ecosystems, human life, and socioeconomic infrastructure\cite{Yang2024}\cite{QIU2025}. In recent decades, the increasing frequency, intensity, and spatial extent of wildfire events have been strongly linked to climate change, prolonged droughts and human activities\cite{Li2021}. Timely and accurate wildfire detection is therefore crucial for effective fire monitoring, early warning and emergency response.

Among various wildfire monitoring tasks, active fire detection focuses on identifying actively burning regions at the time of satellite observation. Compared to burned area mapping, which captures the aftermath of fire events, active fire detection provides near-real-time information which is essential for situational awareness and operational decision-making. However, this task remains challenging due to the sparse spatial distribution of active fires, strong class imbalance, atmospheric interference from clouds and smoke\cite{Ramsey2024}, and the variability of fire signatures across different types of land-cover and observation conditions.

Since the 1970s, large-scale active fire detection has increasingly relied on satellite remote sensing due to its wide spatial coverage and frequent revisit cycles\cite{Chuvieco2020}. Sensors working in the visible, near-infrared and infrared spectral ranges can capture radiometric anomalies associated with combustion processes, while thermal and mid-infrared bands are particularly sensitive to emissions from high-temperature fires. Nevertheless, traditional threshold-based fire products often suffer from false positives and omission errors, motivating the emergence of data-driven approaches.

Recently, deep learning has significantly improved active fire detection performance by enabling models to learn complex spectral, spatial, and temporal patterns directly from satellite observations. In particular, convolutional neural networks and spatiotemporal architectures have demonstrated strong potential in leveraging multi-spectral imagery and time-series data for pixel-level fire detection. The availability of benchmark datasets such as SEN2FIRE and TS-SatFire further facilitates systematic evaluation of deep learning models under diverse geographic conditions, sensor characteristics, and temporal contexts. Building upon these datasets, this work focuses on developing and evaluating deep learning-based methods for robust active fire detection from multi-sensor satellite imagery.

The main contributions of this paper are as follows.
\begin{enumerate}
    \item Successfully build a hybrid network to solve current difficulties and pitfalls of active fire detection with satellite imagery. The result shows the network outperforms several SOTA models in IoU and $F_1$ score. Hence, this network can be considered a baseline for active fire detection.
    \item Research on the function of the spectral attention, channel-spatial modulator, and morphological gates, which we are introducing, and discuss the independent effect and the combined effect of them.
    \item Evaluate the proposed network on two different datasets to demonstrate its novelty and generality.
\end{enumerate}

\section{Related Works}
\subsection{Active Fire Detection}
Active fire detection aims to identify actively burning areas at or near the time of satellite observation by exploiting the radiometric and thermal anomalies induced by combustion processes\cite{Wooster2021}. Early operational active fire products were primarily based on thresholding techniques applied to mid-infrared and thermal infrared bands, where active fires exhibit strong sub-pixel thermal signals. Representative examples include the MODIS and VIIRS active fire algorithms, which combine contextual tests, background characterization, and confidence scoring to detect thermal anomalies at global scale \cite{Giglio2003}. While these approaches enable continuous real-time monitoring, they are susceptible to false alarms caused by sun glint, hot surfaces, and industrial heat sources, as well as omission errors under cloud cover or heavy smoke.

Recent advances in machine learning and deep learning have enabled data-driven approaches for active fire detection that move beyond fixed thresholds. Convolutional neural networks and semantic segmentation models have demonstrated robustness in this area by jointly learning spectral and spatial patterns from satellite imagery \cite{Ghali2023} \cite{Thangavel2023}\cite{Rashkovetsky2021}. Moreover, the integration of temporal information through satellite image time series has been shown to further enhance detection reliability by capturing fire evolution dynamics and suppressing transient noise \cite{Zhao2022}.

Despite these advances, active fire detection remains a challenging task due to the sparse spatial distribution of fire pixels, severe class imbalance, atmospheric interference from clouds and smoke, and large variability in fire signatures across different land-cover types and observation conditions. These challenges motivate the continued development of robust deep learning models and benchmark datasets tailored specifically for active fire detection under realistic operational scenarios.
\subsection{U-Net Based Models}
U-Net is a fully convolutional encoder--decoder architecture originally developed for biomedical image segmentation, and has since become a standard model for dense prediction tasks in remote sensing \cite{Ronneberger2015}\cite{He2022}. The architecture consists of a contracting path that progressively extracts hierarchical features and an expansive path that restores spatial resolution through upsampling operations, which are also called the encoder and decoder, respectively. Furthermore, skip connections between corresponding encoder and decoder layers enable the preservation of fine-grained spatial details while incorporating high-level contextual information.

Due to its ability to produce accurate pixel-level predictions with limited training data, U-Net has been widely adopted for wildfire-related tasks such as active fire detection and burned area mapping. Numerous studies have demonstrated that U-Net-based models outperform traditional threshold-based methods by learning complex spectral and spatial patterns associated with fire activity\cite{SUI2024}\cite{Cui2023}. Furthermore, U-Net serves as a flexible architectural template that can be enhanced with residual blocks, attention mechanisms\cite{SHLOMO2025}, and temporal modeling components, making it a strong baseline for modern deep learning-based wildfire detection systems.

Residual Attention U-Net (RAUNet), a novel encoder-decoder architecture, was proposed to address overlooking or misclassification in the standard U-Net architecture\cite{Ni2019}. It applies global average pooling (GAP) to extract global context, models channel dependencies, and uses high-level deep semantic data to select and filter low-level spatial details. Rather than concatenating features, it fuses them using element-wise matrix addition, significantly reducing computational overhead and network parameters.

\section{Data}
Before introducing the proposed method, we first review the datasets used in this study. Our proposed model is described in the next subsection. To address the advantages and flexibility of our proposed model, we select 2 different datasets focused on wildfire detection.
\subsection{TS-SatFire}
TS-SatFire is a large-scale, multi-task satellite image time-series dataset developed to support active fire detection, burned area mapping, and wildfire progression prediction\cite{Zhao2025}. It captures the full temporal evolution of wildfire events across the contiguous United States between 2017--2021, encompassing 179 distinct fire events and more than 3{,}500 multi-temporal satellite image samples. Unlike static-image benchmarks, TS-SatFire explicitly models wildfire dynamics by organizing data as temporally ordered sequences.

The dataset is primarily based on VIIRS imagery acquired from multiple polar-orbiting satellites, with a spatial resolution of 375~m for imagery bands and 750~m for moderate bands. Six core spectral bands are selected for fire monitoring, including mid-infrared and long-wave infrared channels that are highly sensitive to active fire signals. In addition to satellite imagery, TS-SatFire integrates a rich set of auxiliary variables, including vegetation indices, meteorological conditions, weather forecasts, topographic attributes, and land-cover information, enabling multi-modal learning for both detection and prediction tasks.

Pixel-level labels for active fires and burned areas are derived from a combination of VIIRS active fire products and official fire perimeter records, supplemented by manual quality control to mitigate false positives and labeling noise. For wildfire progression prediction, labels are defined as the newly burned areas between consecutive days, encouraging models to learn genuine spatiotemporal fire spread rather than static segmentation. In this study, we specify this dataset for active fire detection. So we remove the time series and treat the images as independent samples. Through its flexible design, TS-SatFire provides a comprehensive benchmark for wildfire monitoring applications.
\subsection{Sen2Fire}
The Sen2Fire dataset is a challenging benchmark designed for wildfire detection using multisource Sentinel satellite data\cite{Xu2024}. It is constructed from high-resolution Sentinel-2 multispectral imagery combined with Sentinel-5P aerosol products, and focuses on major bushfire events that occurred during the 2019--2020 Australian wildfire season. The dataset covers four geographically distinct regions in New South Wales, Australia, enabling the evaluation of model robustness and spatial transferability across heterogeneous fire-affected landscapes.

Each sample in Sen2Fire consists of image patches of size $512 \times 512$ pixels with a spatial resolution resampled to 10~m. The input features include 12 spectral bands from Sentinel-2 (spanning visible, near-infrared, and short-wave infrared wavelengths) and one aerosol index channel derived from Sentinel-5P, resulting in 13-channel inputs. Ground-truth wildfire labels are generated from the MOD14A1~V6.1 active fire product from the National Aeronautics and Space Administration (NASA), providing pixel-level binary annotations for fire and non-fire regions. The dataset is partitioned into non-overlapping training, validation, and test sets based on geographical separation to prevent spatial leakage. Notably, Sen2Fire exhibits a strong class imbalance, reflecting the realistic sparsity of fire pixels in large-scale satellite imagery and posing a non-trivial segmentation challenge for deep learning models.

\section{Methods}
We propose a channel-attentioned dynamic-gated encoder--decoder network, termed SMA-UNet, designed specifically for active fire detection from multi-spectral satellite imagery. The model is built on a lightweight U-Net architecture augmented with residual connections, spectral attention, and dynamically learned refinement modules to address the inherent challenges of active fire detection, including sparse foreground distribution, strong class imbalance and large spectral variability across observation conditions.

\begin{figure*}[!t]
\centering
\includegraphics[width=1\linewidth]{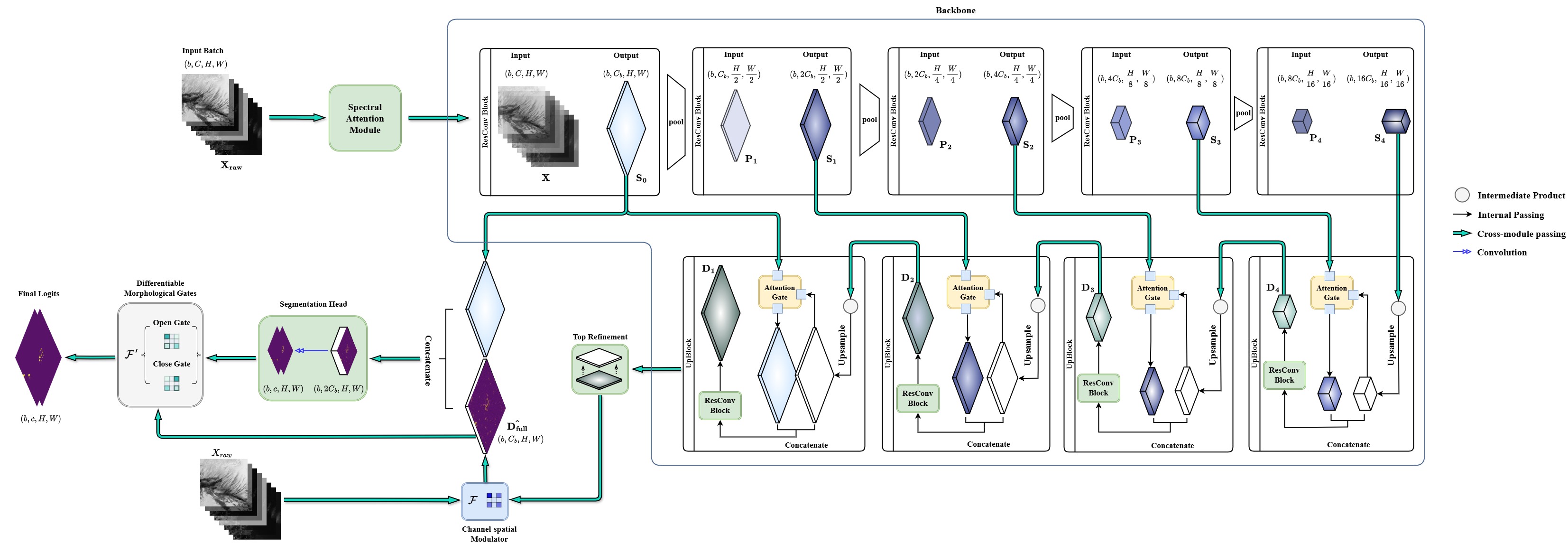}
\caption{Proposed model architecture. To make the visual effect clearer, the figure shows the case where $b=1$. The input $X_{raw}$ is a batch from a dataset, and the output is a $(b,c,H,W)$ foreground and background map, where $c$ is the number of segmented classes. $\mathcal{F}$ and $\mathcal{F^{\prime}}$ represent a series of operations inside the corresponding block.}
\label{ModelArchitecture}
\end{figure*}

\subsection{Spectral Attention Module} 
\begin{figure}
\centering
\includegraphics[width=1\columnwidth]{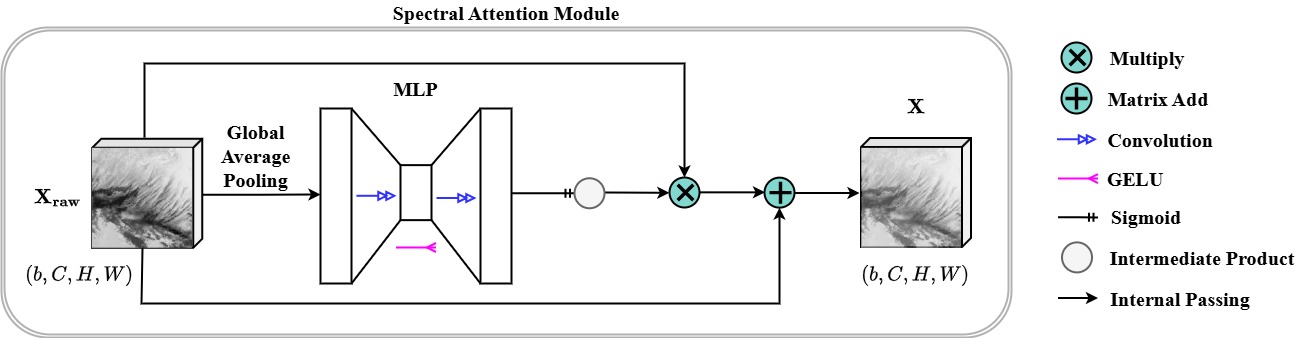}
\caption{Spectral attention module. The input $X_{raw}$ and output $X$ are the same in size. The input passes through an MLP, a sigmoid activation function, and a residual connection.}
\label{SpectralAttentionModule}
\end{figure}

Inspired by the convolutional block attention module (CBAM)\cite{Woo2018}, we implement a spectral attention module (Fig.~\ref{SpectralAttentionModule}) before the input feeds into the UNet-style backbone. This module computes a refined feature representation by re-weighting input spectral channels based on global spatial context. We denote the input tensor as $\mathbf{X_{raw}} \in \mathbb{R}^{B \times C \times H \times W}$, where $B$, $C$, $H$, $W$ are the batch size, input channels, image height, and image width, respectively. The pipeline is defined as squeeze-excitation-residual scaling (SERS)\cite{Hu2018}. The first step is squeeze (global information embedding), which aggregates spatial information into a channel descriptor $\mathbf{z} \in \mathbb{R}^{C_ \times 1 \times 1}$ using global average pooling (GAP): 
\[
\mathbf{z}=\mathbf{GAP}(\mathbf{X_{raw}}).
\]
Then the excitation function block (also called channel dependency modeling) is applied, with a bottleneck Multi-Layer Perceptron (MLP) followed by a Sigmoid activation to produce the attention weights $\mathbf{w}$:
\[
\mathbf{w} = \sigma(\mathbf{MLP}(\mathbf{z}))=\sigma(\mathbf{W}_2 \cdot \phi(\mathbf{W}_1 \cdot \mathbf{z})),
\]
where $\sigma$ denotes the sigmoid function, $\mathbf{W}_1$ and $\mathbf{W}_2$ are $1 \times 1$ convolution layers with a reduction ratio, and $\phi$ denotes the Gaussian error linear unit (GELU) activation function\cite{Hendrycks2023}. 
The output $\mathbf{X}$ is computed by applying the attention weights to the original input and combining it with the identity path through a residual scale factor $\lambda$:
\[
\mathbf{X} = \mathbf{X_{raw}} + \lambda \cdot (\mathbf{X_{raw}} \odot \mathbf{w}),
\]
where $\odot$ denotes element-wise multiplication. This design allows the network to modulate the input by giving different weights to different channels and ensures gradient stability while training.

\subsection{Backbone} 
\begin{figure}
\centering
\includegraphics[width=1\columnwidth]{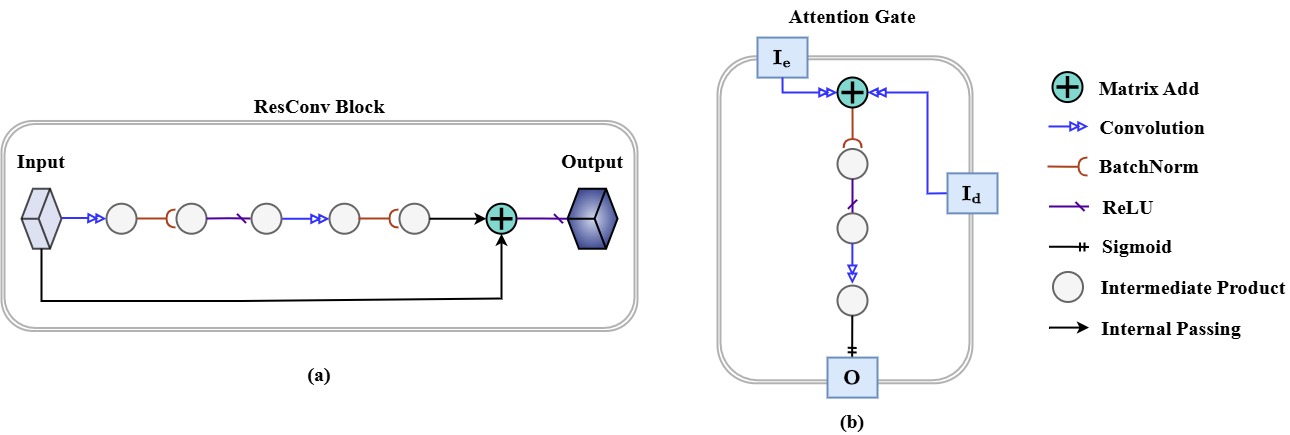}
\caption{Architecture of two components in backbone. (a) ResConv block. (b) Attention Gate. $I_e$ is the low-level features from the downsampler. $I_d$ is the high-level features from the upsampler. $O$ is the output of this block.}
\label{Backbone Components}
\end{figure}

Our architecture utilizes a residual attention UNet style backbone with four downsampling and upsampling stages. At the beginning of the encoder, a special residual convolutional block (ResConv block) is applied to load the reweighted images and resize the output $\mathbf{X}$ into a unified tensor $\mathbf{S_{0}}$ with dimensions $(b,C_b,H,W)$. Each block (shown in Fig.~\ref{Backbone Components}) uses $3\times 3$ convolution as a basic operation, followed by batch normalization and rectified linear unit (ReLU) activation. We repeat these steps and use a residual connection before the final activation. Every downsampling stage outputs an intermediate product $\mathbf{S_{i}}$ with the feature size (the last two dimensions) as the block's input $\mathbf{P_{i}}$, but doubles the channels (the sencond dimension). Between each ResConv block, we use a $2\times2$ max-pooling operation to reduce the feature size of the input fed to the next block. After four downsampling operations, the model outputs a tensor $\mathbf{S_4}$ which has a size of $(b,16C_b,H/16,W/16)$. 
During the upsampling stage, we use skip connections, which concatenate the intermediate products from the attention gate and the intermediate upsampled output. The upsampled output is the product of the first step in every upsample block (marked as UpBlock in Fig.~\ref{ModelArchitecture}). This step uses a transposed convolution with a kernel of $2\times2$ and a stride of 2 to double the spatial dimensions. Then the output is sent to the attention gate, which captures high-level semantic information and the global context in the low-level feature maps. The pipeline goes through a batch normalization, a ReLU function, a $1\times 1$ convolution, and a sigmoid activation function. The output $\mathbf{O}$ is concatenated with the high-level feature map $\mathbf{I_d}$, and sent to the following ResConv block. This step repeats 4 times and outputs a recovered feature map $\mathbf{D_1}$ of $(b,C_b,H,W)$ in dimensions.

\subsection{Top Refinement}
\begin{figure}
\centering
\includegraphics[width=1\columnwidth]{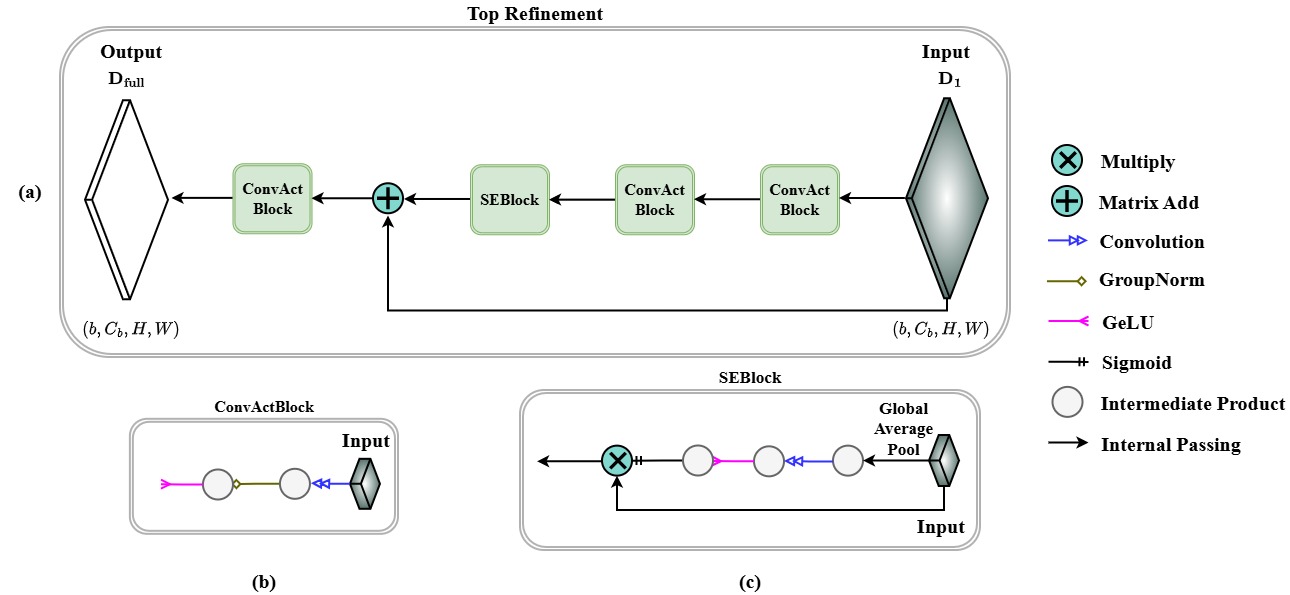}
\caption{Top refinement module and sub-blocks. (a)Top refinement module. The module uses ConvActBlock and SEBlock to calibrate channel attention weights and apply localized spatial smoothing, thereby generates sharpening active fire boundaries $D_{full}$. }
\label{TopRefinement}
\end{figure}

Following the progressive upsampling, we implement a full-resolution refinement stage to harmonize the fused multiscale features. The dimension of the output in the last UpBlock $\mathbf{D_1}$ is consistent with the unified tensor in the first ResConv block, which is $(b,C_b,H,W)$ (shown in Fig.~\ref{TopRefinement}). Then the output is processed through convolution-activation (ConvAct) units and a high-capacity squeeze-and-excitation block (SEBlock). This refinement stage recovers fine-grained spatial details that may have been blurred during upsampling.
SEBlock is applied to enable effective channel-wise feature recalibration. Here, we select group normalization (GN) to replace standard batch normalization (BN), for it performs well when models are trained with small batch sizes \cite{Wu2020}. Similar to the spectral attention module, we choose the GELU as the activation function to handle fire-pixel sparsity in wildfire detection. 

\subsection{Channel-Spatial Modulator} 
\begin{figure}
\centering
\includegraphics[width=1\columnwidth]{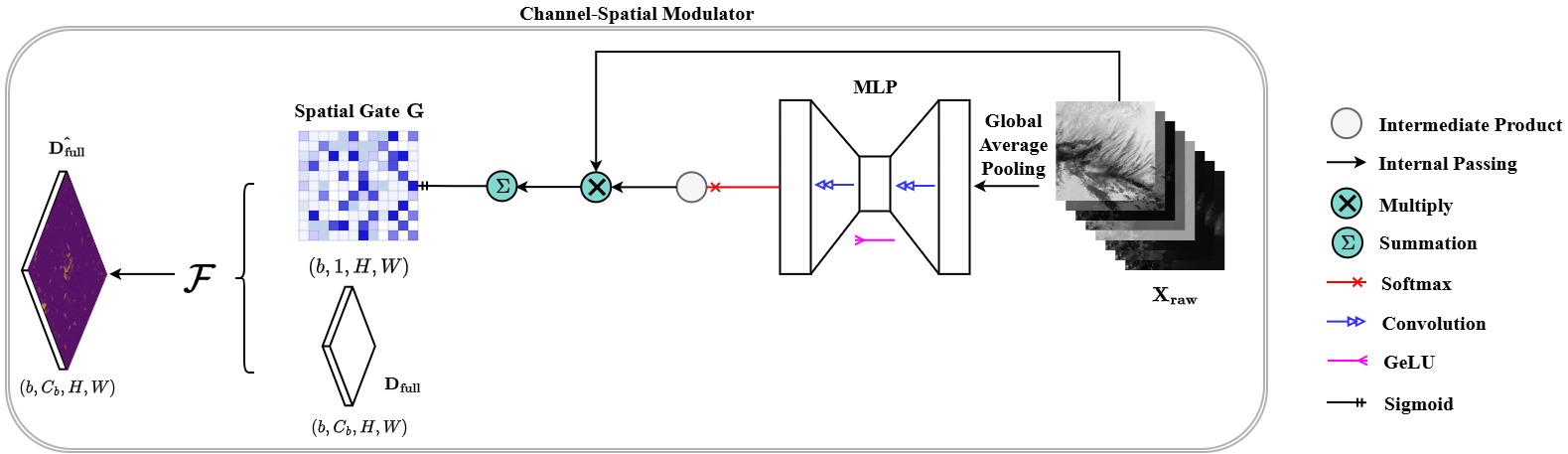}
\caption{Channel-spatial modulator. $\mathcal{F}$ represents the operations applied to spatial gate $G$ and the refined decoder features $D_{full}$.}
\label{FireGate}
\end{figure}

While a SEBlock treats all channels as a weighted vector to scale the features, a top-out strategy is used to compress the entire feature space into a fire-guided map for each batch (Fig.~\ref{FireGate}). Firstly, We compute sample-specific channel importance scores $\mathbf{s} \in \mathbb{R}^{B \times C}$ as same as in the spectral attention module:
\[
s = \mathbf{MLP}(\mathbf{GAP}(\mathbf{X_{raw}})),
\]
The weights over channels can be treated as a probability distribution, thus we apply a Softmax to $\mathbf{s}$. For the $i$-th sample in a batch $B$, the weight $\alpha_{i,j}$ for channel $C_j$ is:
\begin{align*}
\alpha_{i,j} = \frac{\exp(s_{i,j})}{\sum_{k=1}^{C} \exp(s_{i,k})},\\
\sum_{j=1}^{C} \alpha_{i,j} = 1, \quad \forall i \in \{1, \dots, B\}.
\end{align*}
Using these weights, we generate a sample-channel guidance map $\mathbf{M}_i$ by performing a weighted summation across the channel dimension:
$$\mathbf{M}_i = \sum_{j=1}^{C} \alpha_{i,j} \cdot \mathbf{X}_{raw,\{i,j\}}.$$
The corresponding spatial gate $\mathbf{G}_i$ is obtained by passing the guidance map through a refinement network $\mathcal{R}$ and a Sigmoid activation:
$$\mathbf{G}_i = \sigma(\mathcal{R}(\mathbf{M}_i)).$$
The final modulated feature map $\hat{\mathbf{D}}_{full}$ is computed by applying a spatially-varying gain to the refined decoder features $\mathbf{D}_{full}$:
$$\hat{\mathbf{D}}_{full} = \mathbf{D}_{full} \odot \left[ \epsilon + (1 - \epsilon) \cdot \mathbf{G} \right],$$
Where $\epsilon \in [0, 1]$ is a non-negative floor constant that enforces a minimum transparency.

\subsection{Morphological Gates}
\begin{figure}
\centering
\includegraphics[width=1\columnwidth]{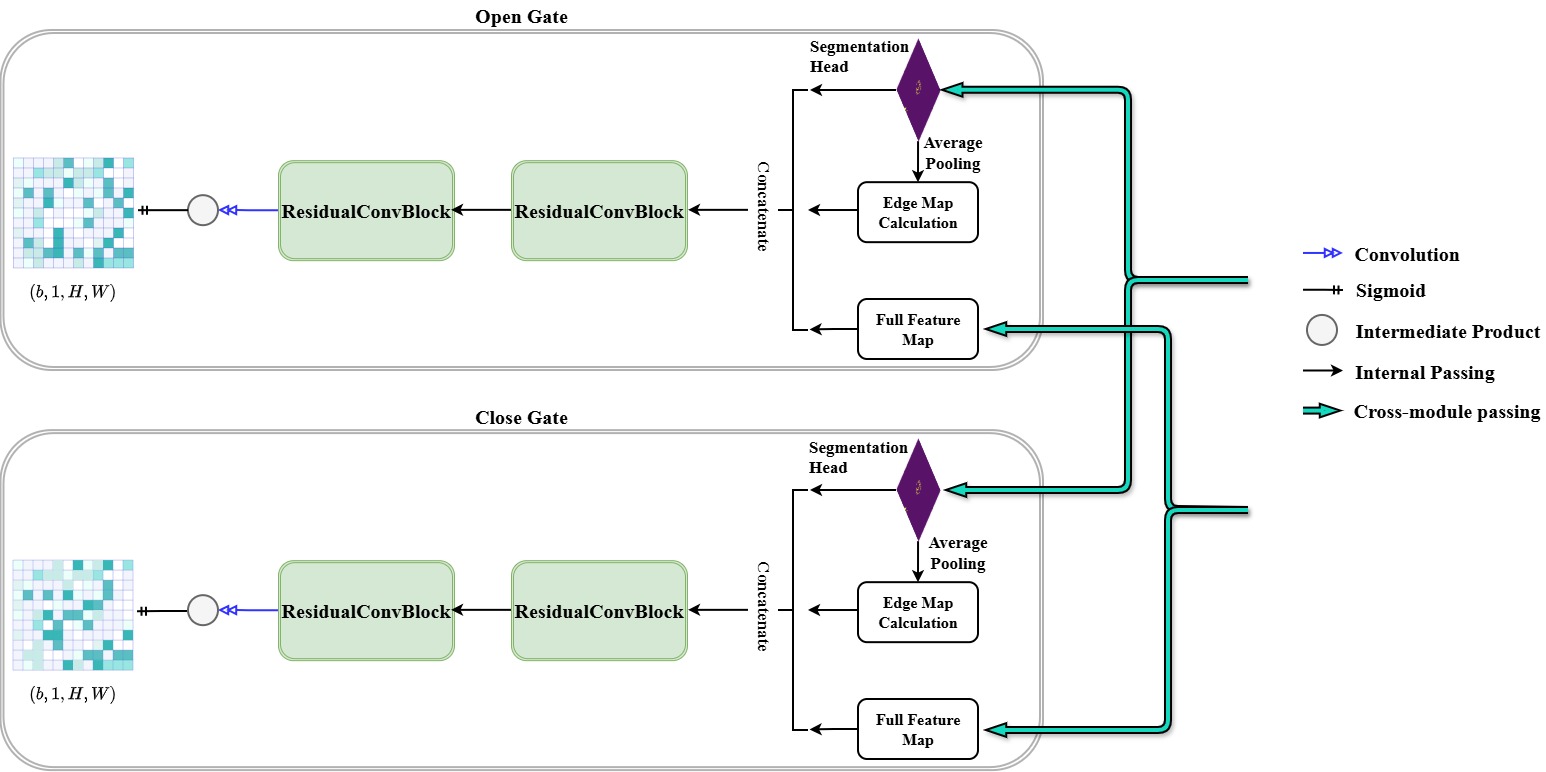}
\caption{A pair of morphological gates, denoted as open gate and close gate.}
\label{BoundaryGates}
\end{figure}

The final module is a pair of boundary-aware gates which consists of an open gate and a close gate (Fig.~\ref{BoundaryGates}). The final prediction logits are produced by a high-resolution fusion head. We concatenate the deep semantic features from the refined decoder ($\mathbf{D}_{full}$) with the low-level spatial features from the initial stem ($\mathbf{S}_0$). This fusion ensures that the classifier has access to both the global context and the finest spatial details of the original image resolution. The fused feature map $\mathbf{X}_{cat}$ is then projected into the class-label space via a convolutional classification head $\mathcal{H}$:
$$\mathbf{L} = \mathcal{H}([\mathbf{D}_{full}, \mathbf{S}_0])$$
Where the logits $\mathbf{L}$ serve as the base prediction for the subsequent morphological gating refinement. Then we convert the initial binary logits into a foreground probability map $\mathbf{P}$ by applying a Softmax operation on $\mathbf{L}$. Furthermore, to intelligently enhance the detection of small fire pixels, we introduce 2 learnable parameters, $\lambda_{open}$ and $\lambda_{close}$, which act as spatially-adaptive mixing coefficients:
$$\lambda_{open} = \sigma(\text{Net}_{open}([\mathbf{P}, \mathbf{D}_{full}, \mathbf{E}])) \cdot \lambda_{max}^{o}$$,
where $\text{Net}_{open}()$ is the application of open gate, $\lambda_{max}^{o}$ is a hyperparameter ceiling to control the intensity of the morphological opening operation, and $\mathbf{E}$ is an edge map from $|\mathbf{P-AP(P)}|$, where $\mathbf{AP}$ represents average pooling.
These coefficients are working with 2 morphological operators, namely open and close, denoted as:
$$\mathbf{P}_{open} = (1 - (1 - \mathbf{P}) * G_{\sigma})* G_{\sigma},$$
$$\mathbf{P}_{close} = 1-(1-\mathbf{P}* G_{\sigma})* G_{\sigma},$$
where $*$ represents a convolution operation, and $G_{\sigma}$ is $3\times3$ Gaussian kernel. So, similar to $\lambda_{open}$, we denote $\lambda_{close}$ as:
$$\lambda_{close} = \sigma(\text{Net}_{close}([\mathbf{P}, \mathbf{D}_{full}, \mathbf{E}])) \cdot \lambda_{max}^{c}$$,
where $\text{Net}_{open}()$ is the application of close gate, $\lambda_{max}^{c}$ is a hyperparameter ceiling to control the intensity of the morphological closing operation.
The final refined probability $\mathbf{P}_{ref}$ is a dynamic fusion:
$$\mathbf{P}_{ref} = (1 - \lambda_{open} - \lambda_{close}) \cdot \mathbf{P} + \lambda_{open} \cdot \mathbf{P}_{open} + \lambda_{close} \cdot \mathbf{P}_{close}$$
where $\lambda_{open}$ and $\lambda_{close}$ are the pixel-wise mixing coefficients predicted by the gating modules. Finally, to map the estimated boundary probability map back to the logit space for multi-class classification, we employ a stabilized logit transform. First, to avoid numerical singularities at the boundaries of the interval $[0, 1]$, the probability map is constrained via:$$\hat{\mathbf{P}}_{{ref}} = \min(\max(\mathbf{P}_{ref}, \epsilon), 1 - \epsilon)$$where $\epsilon = 10^{-6}$ is a small positive regularization constant. The log-odds (logit) of the foreground active fire class is subsequently formulated as:$$\text{logit}_{\text{fg}} = \ln\left(\frac{\hat{\mathbf{P}}_{{ref}}}{1 - \hat{\mathbf{P}}_{{ref}}}\right)$$To align this with standard multi-class categorical cross-entropy frameworks, we construct a 2-channel logit representation by concatenating the implicit background logit ($-\text{logit}_{\text{fg}}$) and the foreground logit.

\subsection{Loss Functions}
The active fire detection is a binary segmentation task, so we choose Dice loss as loss function \cite{Jadon2020}\cite{Mathew2024} in training task of TS-SatFire, written as:
$$L_{Dice} = 1 - \frac{2 \sum_{i=1}^{N} p_i y_i + \epsilon}{\sum_{i=1}^{N} p_i + \sum_{i=1}^{N} y_i + \epsilon},$$
where $y_i$ is the ground truth, $p_i$ is the predicted probability for the foreground class, $N$ is the total number of pixels in the batch, and $\epsilon$ is a very small constant added to prevent division by zero and ensure numerical stability.

In Sen2Fire, a weighted cross-entropy (WCE) Loss is adopted as our loss function to balance the structural sparsity of fire pixels:
$$L_{seg} = -\frac{1}{N} \sum_{i=1}^{N} \left[ w_0 \cdot y_{i,0} \log(p_{i,0}) + w_1 \cdot y_{i,1} \log(p_{i,1}) \right],$$
here, $N$ is the total number of pixels. $w_0 = 1$ and $w_1 = 10$ represent the class weights for the background and active fire classes, respectively. $y_{i,0}$ and $y_{i,1}$ are the one-hot encoded ground truth labels for pixel $i$. $p_{i,0}$ and $p_{i,1}$ are the predicted Softmax probabilities for the background and active fire classes.

\section{Experiments and Discussion}
To demonstrate the performance of our method, we conducted experiments on 2 datasets introduced in the previous section. The experimental results were evaluated with several commonly used metrics. Furthermore, to analyze the function of each component in detail, we made a comprehensive ablation study of the proposed model.
\subsection{Experimental Setup}
Our experiments were performed on Ubuntu 18.04.6, equipped with GeForce RTX 1080 GPUs. We chose PyTorch 2.3.1 with CUDA 11.8 as our basic machine learning platform. As for TS-Satfire, the experiments were completed with a batch size of 4, a learning rate of 0.0003, which is the same as the other compared models. It is worth noting that our model converges more rapidly than many SOTA models, so we choose a total of 30 iterations. In terms of Sen2Fire, we used a batch size of 8, a learning rate of 0.0002, and a total training step of 3000. The input strategy is selecting all channels, because we want to validate our model's performance on dynamically weighting channels. 

We use default setting of UNet and Attention-UNet, which is [64, 128, 256, 512, 1024] in feature size with a stride of 2. In terms of UNetR and Swin-UnetR, we compare them in a similar parameter size, so the implemented feature size is 36 and 48, respectively. For RAUNet, we use a 4-layer encoder and decoder, and the base channel setting is 32.
\subsection{Evaluation Metrics}
\subsubsection{\#param(Number of Parameters)}
This metric indicates the model size and is commonly expressed in units of millions (M). The total number of parameters affects model complexity, memory space, and training speed. Generally, more parameters always lead to higher computational resource consumption. Consequently, in resource-constrained environments, optimizing for parameter efficiency is essential to maintain a balance between predictive performance and deployment feasibility.
\subsubsection{IoU(Intersection over Union)}
IoU, also known as the Jaccard Index, is the most popular metric for object detection and segmentation. IoU is calculated by determining the ratio of the intersection of the predicted and ground truth pixels to the union of those pixels. An outcome where the model correctly predicts the positive class (Fire) is denoted as TP (True Positive). FP represents a score where the model incorrectly predicts the positive class when the actual label is negative, while FN is defined as where the model incorrectly predicts the negative class (Non-fire) when the actual label is positive. The general expression is:
$$IoU  = \frac{TP}{TP + FP + FN},$$
\subsubsection{F1 Score}
In classification contexts, the $F_1$ Score is widely used to measure the harmonic mean of precision and recall. This metric is denoted as:
$$F_{1}~\text{Score} = \frac{2TP}{2TP + FP + FN}$$
Because the $F_1$ Score uses $2 \times TP$ in the numerator, it is slightly more stable, making it a better indicator of whether the model has found the general location and shape of a fire.
\subsection{Comparison with SOTA Models}
\subsubsection{Experiment on TS-SatFire}
Firstly, we compared the proposed architecture with some SOTA models on TS-SatFire. From the results in Table~\ref{TS-SatFire_comparison}, we can find that SMA-UNet achieves the best score in IoU and $F_1$, which are 75.16\% and 84.57\%, respectively. The model uses a relatively small number of parameters, which is only 8.32 M, but significantly outperforms some larger models. 
\begin{table}[htbp]
\caption{Comparison of models on TS-SatFire}
\label{TS-SatFire_comparison}
\centering
\small
\renewcommand{\arraystretch}{1.5}
\setlength\tabcolsep{2pt} 
\begin{tabular}{%
L{0.24\columnwidth}
L{0.24\columnwidth}
L{0.23\columnwidth}
L{0.23\columnwidth}}
\hline
\textbf{Model} &\textbf{\#param(M)} & \textbf{IoU(\%)} & \textbf{$F_1$(\%))} \\
\hline
U-Net & 10.55 & 62.89 & 73.52\\
\hline
Attention-UNet & 31.74 & 72.34 & 81.77\\
\hline
SwinUNetR & 25.15 & 72.71  & 82.10 \\
\hline
RAUNet & \textbf{8.20} & 72.78 & 82.18 \\
\hline
SMA-UNet & 8.32 & \textbf{75.16} & \textbf{84.57} \\
\hline
\multicolumn{4}{@{}p{0.94\columnwidth}@{}}{\footnotesize \textit{Note:} \textbf{bold} represents the best.}
\end{tabular}
\end{table}

\begin{figure}
\centering
\includegraphics[width=0.75\columnwidth]{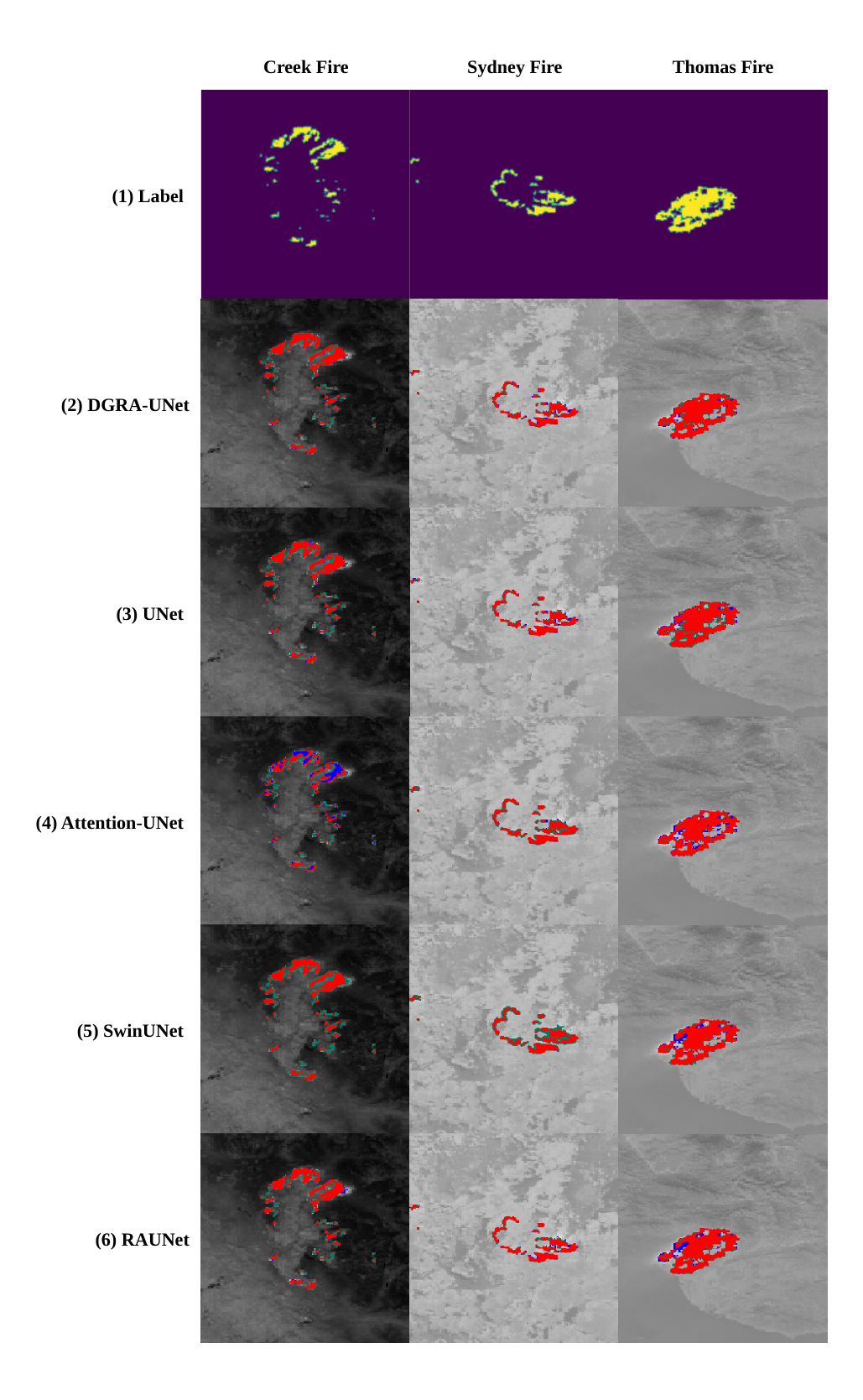}
\caption{A comparison of test results in TS-SatFire. We select representative samples corresponding to 3 different sites. Images in row (1) are the ground truth labels of fire mask. Rows (2) to (6) are the detection results of SMA-UNet, UNet, Attention-UNet, SwinUNet, and RAUNet, respectively. The red area represents the correct detections, while the green and blue pixels represent the false alarms and missing fires, respectively.}
\label{Fig Comparison on TS-SatFire}
\end{figure}

We also use exemplary samples to visualize the test results of all models. We use different colors to mark the detected results. As shown in Fig~\ref{Fig Comparison on TS-SatFire}, the proposed model SMA-UNet perform better than other models, for it have more red area with fewer blue pixels. Furthermore, SMA-UNet has a clear advantage at the edge of the fire source, especially around the perforated patterns. Obvious in the Creek Fire and Sydney Fire, although UNet and SwinUNet can cover more fire pixels, their ability to constrain at the edges is weak, leading to a lot of false alarms. Overall, compared with other models, SMA-UNet constrains the boundaries much better, which means stronger capabilities in contour representation and detail recognition. 

\subsubsection{Experiment on Sen2Fire}
To demonstrate the scalability and generalization ability of the model, we compared the proposed architecture with some SOTA models on Sen2Fire. As shown in Table~\ref{Sa2tFire_comparison}, SMA-UNet has achieved the highest scores, which is 22.50\% in IoU and 36.73\% in $F_1$.

\begin{table}[htbp]
\caption{Comparison of models on Sen2Fire (\textbf{bold} represents the best)}
\label{Sa2tFire_comparison}
\centering
\small
\renewcommand{\arraystretch}{1.5}
\setlength\tabcolsep{2pt} 
\begin{tabular}{%
L{0.24\columnwidth}
L{0.24\columnwidth}
L{0.23\columnwidth}
L{0.23\columnwidth}}
\hline
\textbf{Model} &\textbf{\#param(M)} & \textbf{IoU(\%)} & \textbf{$F_1$(\%))} \\
\hline
U-Net & 10.55 & 20.65 & 34.23 \\
\hline
Attention-UNet & 31.75 & 18.61 & 31.38\\
\hline
SwinUNetR & 25.16 & 4.08 & 7.85 \\
\hline
RAUNet & \textbf{8.20} & 7.61 & 14.14\\
\hline
SMA-UNet & 8.32 & \textbf{22.50} & \textbf{36.73} \\
\hline
\multicolumn{4}{@{}p{0.94\columnwidth}@{}}{\footnotesize \textit{Note:} \textbf{bold} represents the best.}
\end{tabular}
\end{table}

\begin{figure}
\centering
\includegraphics[width=0.7\columnwidth]{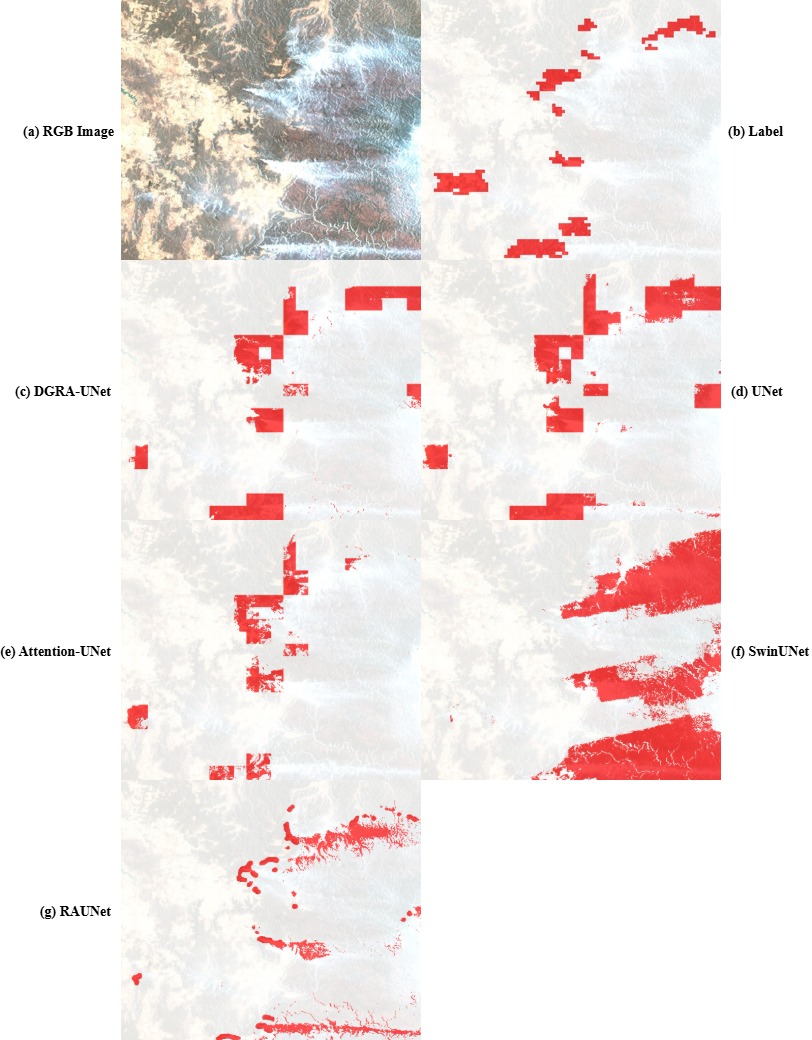}
\caption{A comparison of test results in Sen2Fire. The red pixels represent fire points inferred by the model.}
\label{Fig Comparison on Sen2Fire}
\end{figure}

Similar to TS-Satfire, we demonstrate the visualization of the test results on Sen2Fire. The test set in Sen2Fire consists of 504 patches, but all of them are from the same area in Australia. The visualization merges these patches into a single image. The test images of Sen2Fire contain not only fire points, but also a lot of smoke. This test set can evaluate the model's ability to extract key information from remote sensing images with many interference factors. The results in Fig~\ref{Fig Comparison on Sen2Fire} depict that the SMA-UNet is closer to the ground truth label, with a stronger ability to identify fire source pixels. It is worth noting that RAUNet is almost impossible to distinguish between smoke and the fire source, which ultimately leads to a very low IoU of 7.61\%.

\subsection{Ablation Study}
\begin{table}[htbp]
\caption{Ablation study results of modules on TS-SatFire}
\label{TS-SatFire_ablation}
\centering
\small
\renewcommand{\arraystretch}{1.5}
\setlength\tabcolsep{2pt} 
\begin{tabular}{%
L{0.24\columnwidth}
L{0.24\columnwidth}
L{0.23\columnwidth}
L{0.23\columnwidth}}
\hline
\textbf{Module Implemented} &\textbf{\#param(M)} & \textbf{IoU(\%)} & \textbf{$F_1$(\%)} \\
\hline
Full & 8.32 & 75.16 & 84.57\\
\hline
None (Baseline) & 8.23 & 72.29 & 81.95\\
\hline
+CG & 8.27 & 72.74 & 82.42\\
\hline
+OG & 8.27 & 73.03 & 82.36 \\
\hline
+OG +CG & 8.31 & 73.22 & 82.79\\
\hline
+CSM & 8.24 & 72.79 & 82.69\\
\hline
+CSM +CG & 8.28 & 73.39 & 83.13 \\
\hline
+CSM +OG & 8.28 & 73.61 & 83.00\\
\hline
+CSM +OG +CG & 8.32 & 74.00 & 83.28 \\
\hline
+SA  & 8.23 & 73.13 & 82.77 \\
\hline
+SA +CG & 8.27 & 73.38 & 82.98\\
\hline
+SA +OG & 8.27 & 73.26 & 82.81 \\
\hline
+SA +OG +CG & 8.31 & 73.63 & 83.12\\
\hline
+SA +CSM & 8.24 & 73.68 & 83.08 \\
\hline
+SA +CSM +CG & 8.28 & 74.19 & 83.49\\
\hline
+SA +CSM +OG & 8.28 & 74.18 & 83.50\\
\hline
\multicolumn{4}{@{}p{0.94\columnwidth}@{}}{\footnotesize \textit{Note:} +CG, +OG, +CSM, and +SA represent models implemented with the close gate, open gate, channel-spatial modulator, and spectral attention, respectively.}
\end{tabular}
\end{table}
To discuss the contribution of each component in the proposed model, we conducted comprehensive ablation studies on both datasets.
We enumerate the complete set of module combinations and list the ablation result of TS-SatFire in Table~\ref{TS-SatFire_ablation}. The result shows that each module provides a different improvement in detection performance. These modules are small in size, especially the spectral attention module, but all of them have enhancements in IoU and $F_1$. The fully implemented model improves IoU by 2.87\%, and 2.62\% in $F_1$. In terms of each independent module, the spectral attention module improves the detection performance the most, which achieves a score of 73.13\% in IoU and 82.77\% in $F_1$. Combining these modules can further improve detection performance, e.g., the open and close gate combination gets a higher IoU (73.22\%) than the separate implementations (73.03\% and 72.74\%). Furthermore, sets with 3 modules outperform those with 2 modules in this dataset.

\begin{figure}
\centering
\includegraphics[width=0.45\columnwidth]{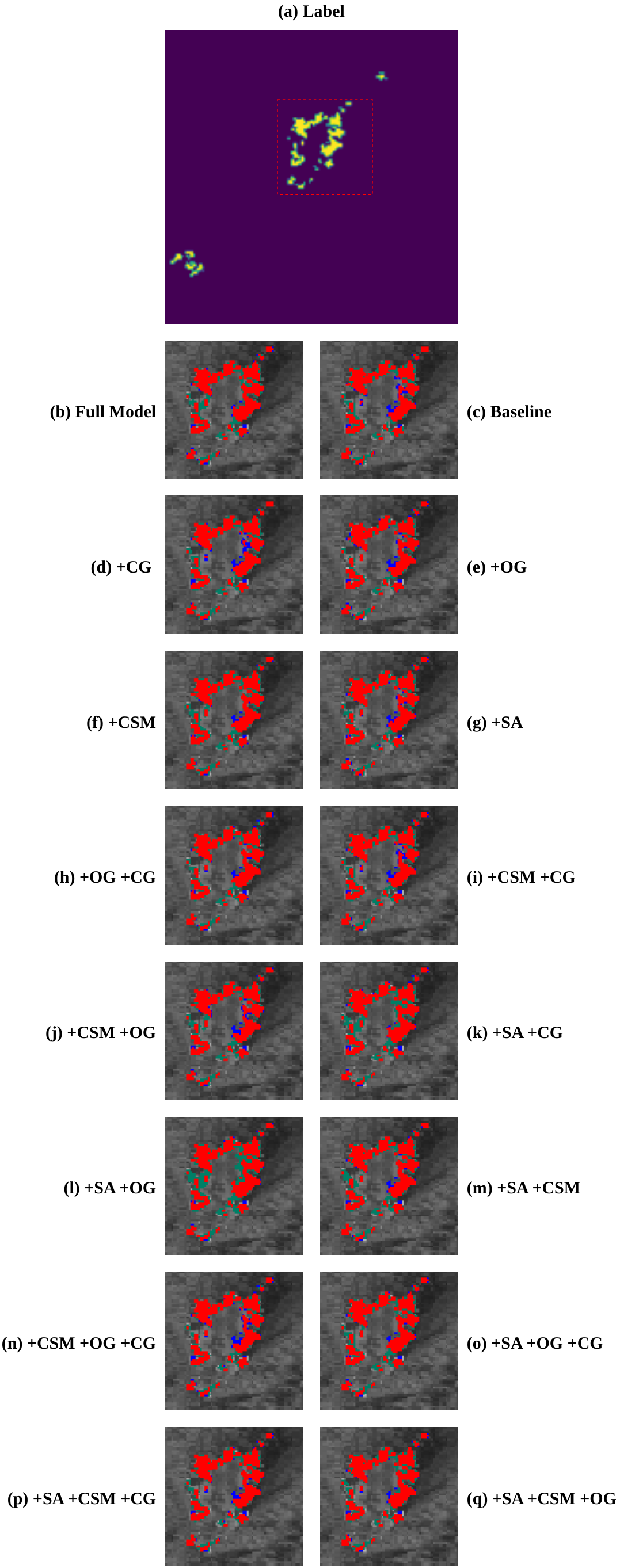}
\caption{A fire sample in TS-SatFire (location: Double Creek). (a) Ground truth label (red dashed box is the comparison area). (b) Full implemented model. (c) Only implemented backbone. (d)-(g) are models implemented with one module. +CG, +OG, +CSM, and +SA represent models implemented with the close gate, open gate, channel-spatial modulator, and spectral attention, respectively. (h)-(m) are models implemented with 2 modules. (n)-(q) are models implemented with 3 modules.}
\label{Fig Ablation on TS-SatFire}
\end{figure}
To more intuitively demonstrate the effects of each module, we selected some results from the test set for comparison. Fig.~\ref{Fig Ablation on TS-SatFire} shows the difference between the detection results of these four modules. From the subgraphs (d) to (g), which highlight the effect of these modules by independently adding each of them, we can find significant differences in improving detection results. As expected, the open operation is better for pattern recognition when multiple ignition points are connected, while the closed operation is the opposite. The fire gate can enhance the detection result by covering more detected pixels, but has a higher false alarm rate. Frame has a lower false alarm rate but misses more fires at the boundaries. Subgraphs (f) to (m) illustrate the performance of each combination of these modules. Same as in the Table~\ref{TS-SatFire_ablation}, the combination of 3 modules works better than combining 2 modules or using a single module. 
\begin{table}[htbp]
\caption{Ablation study results of modules on Sen2Fire}
\label{Sen2Fire_ablation}
\centering
\small
\renewcommand{\arraystretch}{1.5}
\setlength\tabcolsep{2pt} 
\begin{tabular}{%
L{0.24\columnwidth}
L{0.24\columnwidth}
L{0.23\columnwidth}
L{0.23\columnwidth}}
\hline
\textbf{Module} &\textbf{\#param(M)} & \textbf{IoU(\%)} & \textbf{$F_1$(\%)} \\
\hline
Full & 8.32 & 22.50 & 36.73\\
\hline
None (Baseline) & 8.23 & 16.29 & 28.01\\
\hline
+CG & 8.27 & 18.12 & 30.69\\
\hline
+OG & 8.27 & 16.41 & 28.19\\
\hline
+OG +CG & 8.31 & 19.50 & 32.64\\
\hline
+CSM & 8.24 & 15.69 & 27.13\\
\hline
+CSM +CG & 8.28 & 15.68 & 27.10 \\
\hline
+CSM +OG & 8.28 & 17.13 & 29.25\\
\hline
+CSM +OG +CG & 8.32 & 19.06 & 32.02 \\
\hline
+SA  & 8.23 & 16.47 & 28.29 \\
\hline
+SA +CG & 8.27 & 19.11 & 32.09 \\
\hline
+SA +OG & 8.27 & 18.78 & 31.62 \\
\hline
+SA +OG +CG & 8.31 & 18.89 & 31.78 \\
\hline
+SA +CSM & 8.24 & 20.81 & 34.45 \\
\hline
+SA +CSM +CG & 8.28 & 13.79 & 24.24\\
\hline
+SA +CSM +OG & 8.28 & 12.06 & 21.52\\
\hline
\multicolumn{4}{@{}p{0.94\columnwidth}@{}}{\footnotesize \textit{Note:} +CG, +OG, +CSM, and +SA represent models implemented with the close gate, open gate, channel-spatial modulator, and spectral attention, respectively.}
\end{tabular}
\end{table}

\begin{figure}
\centering
\includegraphics[width=0.6\columnwidth]{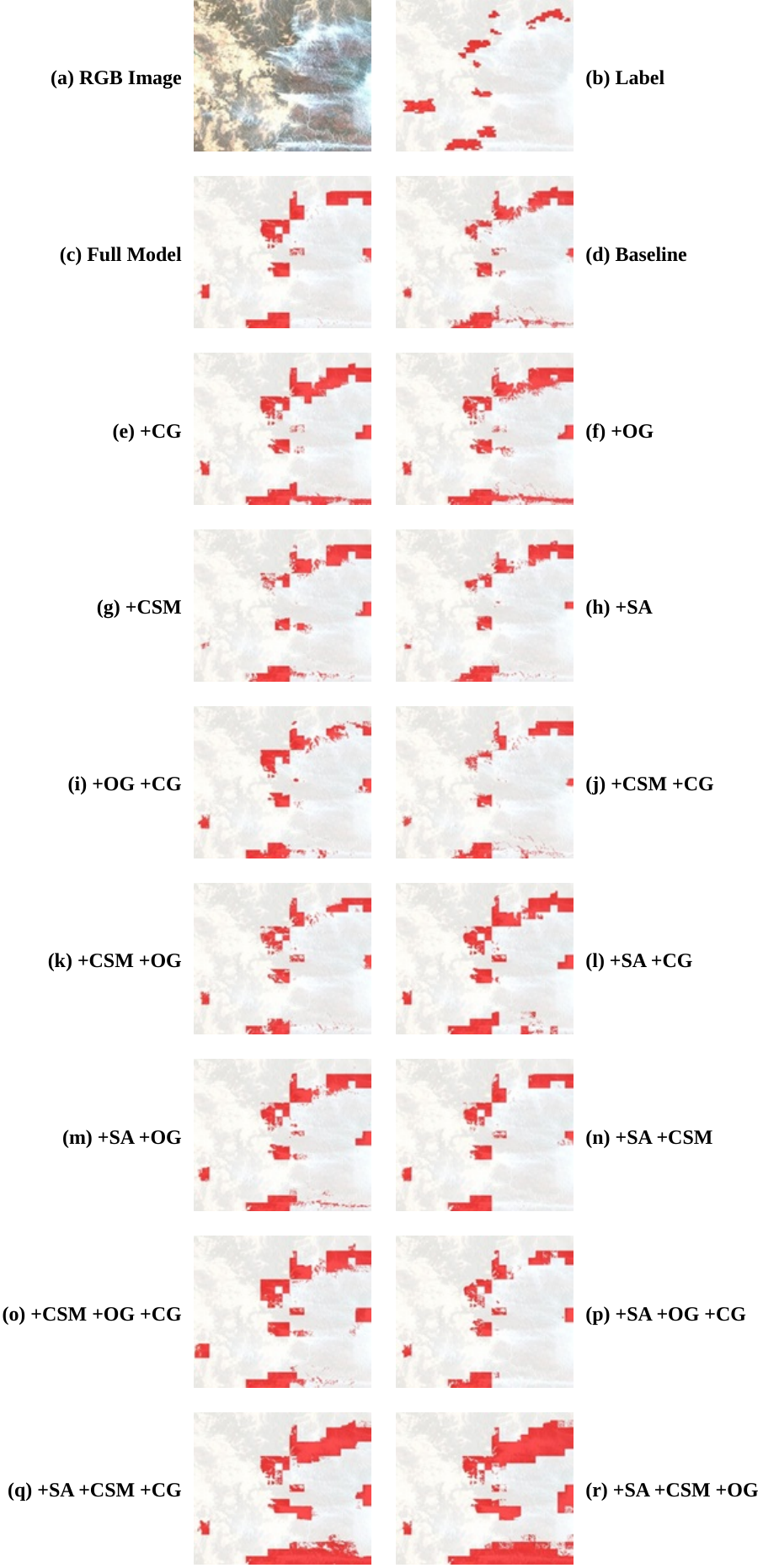}
\caption{Visual results of ablation studies in Sen2Fire. (a) Test image in visible band (RGB). (b) Ground truth label (red area represents wildfire). (c) Full implemented model. (d) Only implemented backbone. (e)-(h) are models implemented with one module. +CG, +OG, +CSM, and +SA represent models implemented with the close gate, open gate, channel-spatial modulator, and spectral attention, respectively. (i)-(n) are models implemented with 2 modules. (o)-(r) are models implemented with 3 modules.}
\label{Fig Ablation on Sen2Fire}
\end{figure}
We also compared the performance of these modules in the Sen2Fire dataset. Since the model is adaptive to the input channels and keeps the same backbone, the model sizes in Table~\ref{Sen2Fire_ablation} are close to the numbers in Table~\ref{TS-SatFire_ablation}. The concise and efficient modules still apparently improve the model's performance on Sen2Fire. The fully implemented model achieved 22.50\% in IoU and 36.73\% in $F_1$, which is 6.21\% and 8.72\% higher than the baseline model. Apparently, the open and close gates still enhance the final detection in this dataset. However, when only implementing the fire gate module on the baseline, the result shows that this module hinders the final performance slightly. As shown in Figure~\ref{Fig Ablation on Sen2Fire}, although the combination of spectral attention and fire gate modules can detect more fire spots, it also significantly increases false alarms. In the end, relying on dynamic open and close operations, the full framework efficiently improves the final score by balancing false positives and false negatives.

\section{Conclusion}
In this paper, we proposed a new model named SMA-UNet, which includes a UNet backbone, a spectral attention module, a fire gate, and a pair of boundary gates. We evaluated this model with 2 datasets, i.e., TS-SatFire and Sen2Fire. In TS-SatFire, SMA-UNet achieved a score of 75.16\% in IoU, and 84.57\% in $F_1$. In Sen2Fire, SMA-UNet achieved a score of 22.50\% in IoU, and 36.73\% in $F_1$. The scores achieved in both datasets are higher than those of other SOTA models. We also conducted ablation studies on both datasets and found that every module has its own independent contribution to the completed model. In future endeavors, we plan to expand our model to more scenarios and reduce the false detection rate based on the current detection level.

\clearpage  

\bibliographystyle{unsrtnat}
\bibliography{references}

\end{document}